\documentclass[10pt, a4paper]{article}

\usepackage[final]{lrec2026} 

\usepackage{microtype}

\usepackage{inconsolata}

\usepackage{graphicx}
\usepackage{CJKutf8}
\usepackage{listings}
\usepackage[most]{tcolorbox}

\usepackage[linesnumbered,ruled,vlined]{algorithm2e}
\usepackage{multirow} 
\usepackage{amsmath} 
\usepackage{booktabs}
\lstdefinelanguage{json}{
    basicstyle=\ttfamily\small,
    numbers=left,
    numberstyle=\tiny\color{gray},
    stepnumber=1,
    numbersep=5pt,
    showstringspaces=false,
    breaklines=true,
    frame=single,
    backgroundcolor=\color{lightgray},
    morestring=[b]',
    morecomment=[l]{//}
}

\definecolor{gray}{rgb}{0.5,0.5,0.5}
\definecolor{lightgray}{rgb}{0.95,0.95,0.95}
\definecolor{MyBlue}{RGB}{137, 207, 240}

\usepackage{xcolor}
\usepackage{colortbl}
\usepackage{pgf}

\newcommand{\cval}[1]{%
    \pgfmathsetmacro{\val}{#1}%
    \definecolor{heatcolor}{rgb}{1,1,1}%
    \pgfmathparse{100*(1 - \val)} 
    \xdef\myval{\pgfmathresult}%
    \cellcolor{MyBlue!\myval}%
}

\usepackage{cleveref}

\title{Assessing the Effectiveness of LLMs in Delivering\\Cognitive Behavioral Therapy}

\name{Navdeep Singh Bedi$^{1}$, Ana-Maria Bucur$^{1}$, Noriko Kando$^{2}$ and Fabio Crestani$^{1}$} 

\address{$^{1}$Università della Svizzera italiana, Switzerland\\
         $^{2}$National Institute of Informatics, Japan\\
         \{ana-maria.bucur-cosma, fabio.crestani\}@usi.ch\\}

\abstract{
As mental health issues continue to rise globally, there is an increasing demand for accessible and scalable therapeutic solutions. Many individuals currently seek support from Large Language Models (LLMs), even though these models have not been validated for use in counseling services. In this paper, we evaluate LLMs' ability to emulate professional therapists practicing Cognitive Behavioral Therapy (CBT). Using anonymized, transcribed role-play sessions between licensed therapists and clients, we compare two approaches: (1) a generation-only method and (2) a Retrieval-Augmented Generation (RAG) approach using CBT guidelines. We evaluate both proprietary and open-source models for linguistic quality, semantic coherence, and therapeutic fidelity using standard natural language generation (NLG) metrics, natural language inference (NLI), and automated scoring for skills assessment. Our results indicate that while LLMs can generate CBT-like dialogues, they are limited in their ability to convey empathy and maintain consistency. 
\\\newline  }

\begin{document}

\maketitleabstract

\section{Introduction and Related Work}
Mental health disorders\makeatletter{\renewcommand*{\@makefnmark}{}
\footnotetext{\color{red}Disclaimer: this study does not support or endorse the use of LLMs in therapeutic practice.}\makeatother} are increasingly prevalent, highlighting a significant need for accessible digital interventions to compensate for the shortage of psychologists and counselors capable of providing therapeutic support.\footnote{\href{https://www.who.int/news/item/17-06-2022-who-highlights-urgent-need-to-transform-mental-health-and-mental-health-care}{https://www.who.int/news/item/17-06-2022-who-highlights-urgent-need-to-transform-mental-health-and-mental-health-care}} As individuals become more inclined to express their concerns through online tools rather than engaging in face-to-face therapy \cite{lucas2014s,haensch2025listens}, there is growing interest in the potential of generative AI chatbots for facilitating the delivering of therapy. Although recent studies have reported a positive shift in attitudes towards using Large Language Models (LLMs) for mental health support \cite{haensch2025listens}, many of these models were not specifically designed for therapeutic applications\footnote{\href{https://thehemingwayreport.beehiiv.com/p/research-report-mental-health-in-an-ai-world}{https://thehemingwayreport.beehiiv.com/p/research-report-mental-health-in-an-ai-world}} and have not been validated for offering counseling services.

The development of dialog systems for emotional support has also gained attention, with researchers focusing on creating frameworks that enable these systems to provide empathetic and effective interaction \cite{liu2021towards,kang2024can}. These systems aim to offer immediate, accessible support, serving as a complementary resource alongside traditional therapeutic practices. However, previous studies have also shown that LLMs are prone to expressing stigma towards people with mental health conditions \cite{moore2025expressing}, and that they respond inappropriately in certain situations, and encourage clients' delusional thinking, most likely due to their training, which tends to promote high agreeableness \cite{hartley-etal-2025-personality}.

With the increasing interest in leveraging technology to enhance psychological counseling \cite{na2025survey}, much of the existing literature has focused on specific therapeutic tasks, such as cognitive reframing techniques. These techniques aim to address cognitive distortions and promote a shift from negative thought patterns to improve mental health outcomes \cite{de2021formulating,chen2023empowering,sharma2023cognitive}. Previous studies have also presented established benchmarks to assess LLM performance in Cognitive Behavioral Therapy (CBT), examining factors such as knowledge acquisition, classification of cognitive distortions and beliefs, and the generation of therapeutic responses \cite{zhang2024cbt}. The findings indicate that while these models exhibit knowledge of cognitive processes, particularly in question-answering scenarios, they generally do not perform at the same level as mental health experts in most generative tasks. Few efforts have investigated the use of LLMs specifically for delivering therapy \cite{lee2024cactus,na2024cbt}. However, most studies rely on synthetic datasets \cite{lee2024cactus,qiu2024smile,de2025introducing,mandal2025magnet,abbasi2025hamraz} rather than authentic therapist dialogues \cite{puspo-etal-2026-mental}, leaving open questions about linguistic fidelity and therapeutic alignment. 

To address this gap, our paper proposes a comprehensive evaluation of LLMs in the context of CBT, a widely recognized and effective form of psychotherapy for a range of mental health conditions, including depression, anxiety, obsessive-compulsive disorder, and various forms of addiction \cite{david2018cognitive}. We compare two distinct methodologies for providing CBT via LLMs: one that leverages existing CBT knowledge without external augmentation, and another that employs a Retrieval-Augmented Generation (RAG) approach. This dual approach enables us to evaluate the extent to which LLMs can generate responses that resemble those of professional therapists in multi-turn dialogues. While previous studies have generated synthetic CBT sessions, the multi-turn interactions in these sessions are generally shorter than those observed in real-life counseling \cite{lee2024cactus}. To address this issue, we use data from real counseling sessions conducted by professional therapists. Our goal is to evaluate how closely utterances generated by LLMs resemble those of professional CBT therapists in multi-turn dialogues. 

The contributions of our study include a thorough comparison of generation-only and RAG methods and a comprehensive analysis of generated responses alongside responses from mental health professionals. Our analysis uses various metrics to evaluate both natural language generation (NLG) and therapeutic effectiveness. This includes assessments for natural language generation, natural language inference (NLI), and automated scoring for general counseling and CBT-specific skills assessment.

The paper aims to answer the following research questions:

\paragraph{RQ1} To what extent can LLMs linguistically and semantically emulate professional therapists in CBT sessions, as measured by natural language generation and inference metrics?

\paragraph{RQ2} Does RAG improve the linguistic quality, semantic coherence, or therapeutic fidelity of LLM-generated CBT responses compared to generation-only methods?

\paragraph{RQ3} How effectively do LLMs demonstrate general counseling skills (e.g., understanding, interpersonal effectiveness, and collaboration), CBT-specific competencies (e.g., guided discovery, focus, and strategy) and empathy relative to human therapists?

Through this exploration, we aim to provide a deeper understanding of LLMs' capabilities in delivering CBT, as many individuals increasingly rely on these models for mental health support. We aim to contribute to a clearer understanding of LLMs' potential and limitations in mental health care.

\section{Methodology}
We evaluate two methods for delivering CBT based on LLMs: (1) a generation-only approach and (2) retrieval-augmented generation approach. Both methods are assessed using transcribed role-play sessions between professional therapists and clients.

\subsection{Data Collection}

\begin{table}[h!]
\centering
\begin{tabular}{l r}
\toprule
\textbf{Metric} & \textbf{Value} \\
\midrule

\multicolumn{2}{l}{\textbf{Client--Therapist Sessions}} \\
\addlinespace
Total sessions & 17 \\
Avg. conversation length & 39 turns \\
Max. conversation length & 74 turns \\
Min. conversation length & 13 turns \\
Avg. utterance length & 28.97 words \\
Max. utterance length & 464 words \\
Min. utterance length & 1 word \\

\addlinespace
\midrule
\addlinespace

\multicolumn{2}{l}{\textbf{Guideline Sessions}} \\
\addlinespace
Total sessions & 26 \\
Avg. length & 127.11 words \\

\bottomrule
\end{tabular}
\caption{Summary statistics of the collected data for the client-therapist sessions and the guidelines sessions.}
\label{dataset-stats}
\end{table}

Because authentic therapy transcripts are rarely public due to privacy constraints, we collected 17 role-played CBT sessions from publicly available YouTube channels run by licensed professionals. Clients were portrayed by actors, ensuring no real patient data was used. We aimed to gather a comprehensive set of video-based therapy sessions, meticulously reviewing each video to extract conversational data. In addition, 26 therapist tutorial sessions explaining CBT principles were collected for use as retrieval guidelines.

Transkriptor\footnote{\href{https://transkriptor.com/}{https://transkriptor.com/}}  was used to transcribe the therapy sessions. After transcription, each transcript was evaluated by two annotators who compared the text with the corresponding audio recordings. While most transcripts were accurate, minor discrepancies were corrected through mutual agreement between the annotators. In addition, the dataset underwent thorough preprocessing and anonymization to remove any personally identifiable information. Finally, the data was converted into a TREC-iKAT.json-like structure \cite{aliannejadi2024trec} for seamless integration into the RAG pipeline.

The 17 counseling sessions had an average conversation length of 39 turns, with each utterance averaging about 28.97 words. The shorter guideline sessions had an average length of 127 words and covered various aspects of CBT, including goal setting during sessions, mood checks, and addressing different client behaviors. A detailed summary of the dataset statistics is presented in Table \ref{dataset-stats}.

\subsection{Generation Approaches}

We evaluate two approaches to generate appropriate utterances for the therapy sessions. The first approach is a generation-only method, motivated by previous research showing that LLMs already possess knowledge of CBT techniques \cite{zhang2024cbt}. The second approach is a retrieval-augmented generation approach, which utilizes information from the guideline sessions.

The generation-only approach uses the following prompt:

\begin{minipage}[t]{0.45\textwidth}
\begin{tcolorbox}[colback=gray!10, colframe=gray!80!black, title=\textbf{Prompt}, fonttitle=\bfseries, coltitle=black]
\ttfamily

\small
You are a psychotherapist specialized in Cognitive Behavioral Therapy (CBT). You are in a session with a patient. Each exchange consists of a therapist's statement followed by the patient's response. Your task is to generate the therapist’s next utterance, taking into account the session history. Your response should be concise, direct, supportive, and grounded in CBT principles. Generate only the therapist’s next utterance. Do not include previous dialogue, multiple conversational turns, explanations, contextual information, or any meta-commentary.\newline

\textbf{Conversation}: <conversation\_history>\newline

\end{tcolorbox}
\end{minipage}

The RAG approach uses the CBT guidelines and follows a two-stage pipeline: retrieving the top three relevant CBT guideline passages based on the conversation context, followed by response generation conditioned on both the conversation history and retrieved passages. We provide an overview of the approach in Algorithm \ref{alg:therapist}. Given a conversation history $C$ between the therapist and the client, we first extract a context window $C_{\text{window}}$ containing the last $w$ dialogue turns. This window captures the immediate conversational context relevant to the current interaction. A window size of three dialogue turns was empirically chosen for optimal context balance. The extracted context window is then encoded into a query representation. Let \( G = \{g_1, g_2, \dots, g_{|G|}\} \) denote the CBT guideline corpus. Using dense retrieval, we retrieve the top-$k$ most relevant guideline passages (\( G_{top} \)) with respect to the query. To generate the response, we construct the prompt $P$ by concatenating the (1) task-specific instruction text, (2) the retrieved guideline passages (\( G_{top} \)), and (3) the context window ($C_{\text{window}}$). The prompt $P$ is then provided to the LLM,  which generates a suitable response $R_{\text{gen}}$.

\begin{algorithm}[t]
\caption{Therapist Response Generation via Retrieval-Augmented Prompting}
\label{alg:therapist}

\KwIn{Conversation history $C$, guideline corpus $G$, number of retrieved guidelines $k$, window size $w$}
\KwOut{Generated therapist response $R_{\text{gen}}$}

Extract recent window: $C_{\text{window}} \leftarrow$ last $w$ turns from $C$\;
Encode query from $C_{\text{window}}$: $q \leftarrow \text{Encode}(C_{\text{window}})$\;
Retrieve top-$k$ relevant guidelines: $G_{\text{top}} \leftarrow \text{DenseRetrieve}(q, G, k)$\;
Construct prompt $P$ by concatenating:\;
\Indp
Instructions text\;
Retrieved guidelines $G_{\text{top}}$\;
Context window $C_{\text{window}}$\;
\Indm
Generate response: $R_{\text{gen}} \leftarrow \text{LLM}(P)$\;
\Return $R_{\text{gen}}$\;

\end{algorithm}

We use both proprietary and open-source models for generation, including GPT-4o-mini \cite{menick2024gpt}, LLama3 8B \cite{dubey2024llama}, Mistral 7B \cite{Jiang2023Mistral7}, Gemma 7B \cite{team2024gemma}, and Qwen 7B \cite{bai2023qwen}.

\subsection{Evaluation}

To comprehensively assess LLM performance, we employed three complementary evaluation dimensions: linguistic quality and similarity to the therapist's original responses, semantic consistency, and therapeutic and CBT-specific skills. Each of these dimensions captures a distinct aspect of what constitutes effective therapy.

\paragraph{Text generation metrics} The evaluation of generated responses is typically conducted using various text generation metrics \cite{na2024cbt,kang2024can}. We assess metrics that measure the lexical overlap between generated utterances and the therapists' reference responses, such as BLEU \cite{papineni2002bleu}, ROUGE \cite{lin-2004-rouge}, and METEOR \cite{banerjee2005meteor}. In addition, we use BERTScore \cite{zhangbertscore}, which measures semantic similarity. We also consider Distinct-1 \cite{li2016diversity}, which measures the diversity in language generation tasks. However, these measures have inherent limitations, as they may allow models to produce text that is contextually relevant but differs from the responses provided by therapists.

\paragraph{Semantic consistency}  was assessed using natural language inference techniques to evaluate contradictions and coherence. Building on previous research \cite{chim2024overview,laurer2024less}, we employed a DeBERTa V3 large model that has been fine-tuned for NLI.\footnote{\href{https://huggingface.co/MoritzLaurer/DeBERTa-v3-large-mnli-fever-anli-ling-wanli}{https://huggingface.co/MoritzLaurer/DeBERTa-v3-large-mnli-fever-anli-ling-wanli}} This model is used to determine whether the generated response is consistent with or contradicts the original response given by the therapist. Consistency is defined as the absence of contradiction. Since generated responses may consist of multiple sentences, we adopt the approach used by \citet{chim2024overview} and average the maximum contradiction probabilities of the generated sentences from each utterance.

\paragraph{Therapeutic skills evaluation} was conducted to assess general counseling skills, CBT-specific skills, and empathy. Recent research indicates that evaluating natural language generation using LLMs closely aligns with human evaluations and is increasingly used for this task~\cite{chiang2023can,lee2024cactus,kim-etal-2025-multimodal}. For the evaluation of general and CBT skills, we adopt the methodology outlined by \citet{lee2024cactus}. The authors adapted GPT-4o to conduct evaluations based on the Cognitive Therapy Rating Scale (CTRS)\footnote{\href{https://beckinstitute.org/wp-content/uploads/2021/06/CTRS-Full-Documents.pdf}{https://beckinstitute.org/wp-content/uploads/2021/06/CTRS-Full-Documents.pdf}}, which showed a high correlation between the model's predictions and expert manual evaluations. CTRS is a tool designed to assess the CBT skills of therapists, focusing on specific general and CBT skills. The evaluation described by \citet{lee2024cactus} measures a subset of the CTRS that can be applied to text-based settings. The general skills assessed include understanding (e.g., how accurately the therapist comprehends the patient's situation), interpersonal effectiveness (e.g., the therapist's ability to communicate effectively and maintain a positive, professional relationship with the client), and collaboration (e.g., the therapist's capacity to work effectively with the client). The CBT-specific skills evaluated include guided discovery (e.g., the therapist's ability to explore the client's problems and help them draw their own conclusions), focus (e.g., the therapist's ability to concentrate on cognitions and behaviors relevant to the issue), and strategy (e.g., the therapist's ability to implement a consistent strategy for change using CBT techniques). Each of these categories is assigned a score ranging from 0 to 6, in accordance with the guidelines provided by the CTRS tool. We used the same evaluation methodology for CTRS categories that was previously validated by \citet{lee2024cactus}, using GPT-4o to assess the generated responses.

The final measure in our evaluation focuses on \emph{empathy}, which we consider a vital component of the therapeutic alliance \cite{feller2003importance}. The therapeutic alliance refers to the collaborative bond between the therapist and the client, characterized by a strong emotional connection and a mutual understanding of treatment goals and tasks. This alliance is closely linked to the effectiveness of therapy \cite{feller2003importance}. To evaluate empathy, we use the framework developed by \citet{sharma2020computational}, which specifically assesses empathy using a threefold approach that considers emotional reactions, interpretations, and explorations. Emotional reactions (e.g., \textit{I am sorry you are going through...}) relate to affective empathy and help communicate emotions such as warmth and compassion, which are essential for building the therapeutic relationship. Interpretations (e.g., \textit{I understand how you feel...}) are connected to cognitive empathy, indicating that the therapist comprehends the client's feelings. This understanding further enhances the therapeutic alliance. The last component, explorations (e.g., \textit{Are you feeling...?}), is used to probe the client's feelings further. We evaluate these measures of empathy on a scale of 0 to 2.

\section{Results}
\begin{table*}[t!]
    \centering
    \resizebox{1\linewidth}{!}{
    \begin{tabular}{c|cccccccc}
        \toprule
         \textbf{Model} & \textbf{BLEU$\uparrow$} & \textbf{METEOR$\uparrow$} & \textbf{ROUGE-1$\uparrow$} & \textbf{ROUGE-L$\uparrow$} & \textbf{$F1_{BERT}$$\uparrow$} & \textbf{Mean Consistency$\uparrow$} & \textbf{Max Contradiction $\downarrow$} & \textbf{Distinct-1$\uparrow$} \\
         \specialrule{.1em}{.05em}{.05em} 
         GPT-4o-mini & \cval{0.25}1.21 & \cval{0.23}16.64 & \cval{0.13}16.87 & \cval{0.17}12.11 & \cval{0.10}85.77 & \cval{0.10}93.80 & \cval{0.63}24.64 & \cval{0.18}90.72 \\
         GPT-4o-mini$_{G}$ & \cval{0.20}\textbf{1.31} & \cval{0.18}17.10 & \cval{0.11}\textbf{17.05} & \cval{0.15}\textbf{12.20} & \cval{0.08}\textbf{85.82} & \cval{0.20}92.56 & \cval{0.70}26.46 & \cval{0.21}89.97 \\
         LLama3 8B & \cval{0.31}1.10 & \cval{0.36}15.11 & \cval{0.17}16.04 & \cval{0.26}11.08 & \cval{0.23}85.05 & \cval{0.17}92.98 & \cval{0.50}21.95 & \cval{0.20}90.20 \\
         LLama3 8B$_{G}$ & \cval{0.48}0.84 & \cval{0.49}13.39 & \cval{0.23}14.89 & \cval{0.31}10.59 & \cval{0.28}84.77 & \cval{0.34}91.41 & \cval{0.50}21.94 & \cval{0.19}90.67 \\
         Mistral 7B & \cval{0.55}0.71 & \cval{0.23}16.64 & \cval{0.27}14.31 & \cval{0.44}9.40 & \cval{0.30}84.70 & \cval{0.16}93.01 & \cval{0.86}31.61 & \cval{0.38}87.60 \\
         Mistral 7B$_{G}$ & \cval{0.65}0.54 & \cval{0.21}16.85 & \cval{0.27}14.27 & \cval{0.46}9.23 & \cval{0.40}84.13 & \cval{0.21}92.71 & \cval{1.00}38.02 & \cval{0.60}83.38 \\
         Gemma 7B & \cval{0.87}0.26 & \cval{0.89}10.09 & \cval{0.37}12.47 & \cval{0.32}10.58 & \cval{0.17}85.34 & \cval{0.56}90.55 & \cval{0.36}20.28 & \cval{0.00}\textbf{97.70 }\\
         Gemma 7B$_{G}$ & \cval{0.84}0.29 & \cval{0.92}9.77 & \cval{0.38}12.30 & \cval{0.36}10.14 & \cval{0.18}85.29 & \cval{0.38}91.86 & \cval{0.25}\textbf{17.46} & \cval{0.01}97.67 \\
         Qwen 7B & \cval{0.63}0.57 & \cval{0.17}\textbf{17.14} & \cval{0.34}13.43 & \cval{0.60}8.55 & \cval{0.53}83.88 & \cval{0.00}\textbf{95.17} & \cval{0.91}34.98 & \cval{0.70}82.80 \\
         Qwen 7B$_{G}$ & \cval{0.87}0.26 & \cval{0.30}14.43 & \cval{0.42}11.89 & \cval{0.68}7.76 & \cval{0.58}83.50 & \cval{0.37}91.89 & \cval{1.00}42.30 & \cval{0.52}85.36 \\
        \specialrule{.1em}{.05em}{.05em} 
    \end{tabular}
    }
    \caption{Evaluation of generated utterances using text generation metrics (BLEU, METEOR, ROUGE, BERTScore, Distinct-1) and NLI metrics (Mean Consistency, and Maximum Contradictions).}
    \label{tab:text-generation-metrics}
\end{table*}

\begin{table*}[!ht]
    \centering
    \resizebox{\linewidth}{!}{
    \begin{tabular}{c|ccc|ccc}
        \toprule
        \multirow{2}{4em}{\textbf{Model}} & \multicolumn{3}{c|}{\textbf{General Counseling Skills}} & \multicolumn{3}{c}{\textbf{CBT-specific Skills}}\\
        & \textbf{Understanding} & \textbf{Interpersonal Effectiveness} & \textbf{Collaboration} & \textbf{Guided Discovery} & \textbf{Focus} & \textbf{Strategy}\\
        \specialrule{.1em}{.05em}{.05em} 
        GPT-4o-mini & \cval{0.26}3.29 & \cval{0.29}3.76 & \cval{0.82}2.24 & \cval{0.48}3.06 & \cval{0.78}2.47 & \cval{0.78}2.12 \\
        GPT-4o-mini$_{G}$ & \cval{0.32}3.18 & \cval{0.43}3.53 & \cval{0.55}2.82 & \cval{0.32}3.29 & \cval{0.55}2.82 & \cval{0.78}2.12 \\
        LLama3 8B & \cval{0.61}2.59 & \cval{0.69}2.82 & \cval{0.88}2.12 & \cval{0.60}2.71 & \cval{0.82}2.24 & \cval{0.78}2.12 \\
        LLama3 8B$_{G}$ & \cval{0.97}2.00 & \cval{1.00}2.00 & \cval{0.94}2.00 & \cval{1.00}1.88 & \cval{0.94}2.00 & \cval{0.91}2.00 \\
        Mistral 7B & \cval{0.26}3.29 & \cval{0.35}3.65 & \cval{0.55}2.82 & \cval{0.00}\textbf{3.88} & \cval{0.32}3.06 & \cval{0.43}2.71 \\
        Mistral 7B$_{G}$ & \cval{0.32}3.18 & \cval{0.61}3.29 & \cval{0.76}2.35 & \cval{0.17}3.53 & \cval{0.82}2.24 & \cval{0.78}2.12 \\
        Gemma 7B & \cval{0.82}2.35 & \cval{0.65}2.94 & \cval{0.88}2.12 & \cval{0.78}2.47 & \cval{0.88}2.12 & \cval{1.00}1.88 \\
        Gemma 7B$_{G}$ & \cval{0.97}2.00 & \cval{0.74}2.71 & \cval{0.94}2.00 & \cval{0.94}2.00 & \cval{0.94}2.00 & \cval{1.00}1.76 \\
        Qwen 7B & \cval{0.11}3.53 & \cval{0.29}3.76 & \cval{0.82}2.24 & \cval{0.00}\textbf{3.88} & \cval{0.60}2.71 & \cval{0.69}2.35 \\
        Qwen 7B$_{G}$ & \cval{0.78}2.47 & \cval{0.88}2.47 & \cval{0.94}2.00 & \cval{0.10}3.41 & \cval{0.88}2.12 & \cval{0.91}2.00 \\
        \specialrule{.1em}{.05em}{.05em} 
        Therapist & \cval{0.00}\textbf{3.88} & \cval{0.00}\textbf{4.35} & \cval{0.00}\textbf{3.18} & \cval{0.00}3.41 & \cval{0.00}\textbf{3.53} & \cval{0.00}\textbf{3.29} \\
        \specialrule{.1em}{.05em}{.05em} 
    \end{tabular}
    }
    \caption{Evaluation of general counseling and CBT-specific skills.}
    \label{tab:llm-eval}
\end{table*}

\begin{table*}[hbt]
    \centering
    \resizebox{0.7\linewidth}{!}{
    \begin{tabular}{cccc}
        \hline
         \textbf{Model} & \textbf{Empathetic Reactions} & \textbf{Explorations} & \textbf{Interpretations} \\
         \hline
         GPT-4o-mini & \cval{0.28}0.34 & \cval{0.02}1.94 & \cval{1.00}0.00 \\
         GPT-4o-mini$_{G}$ & \cval{0.24}0.37 & \cval{0.00}1.95 & \cval{1.00}0.00 \\
         LLama3 8B & \cval{0.83}0.07 & \cval{0.03}1.92 & \cval{0.99}0.01 \\
         LLama3 8B$_{G}$ & \cval{0.87}0.05 & \cval{0.01}\textbf{1.96} & \cval{1.00}0.00 \\
         Mistral 7B & \cval{0.10}0.48 & \cval{0.91}0.78 & \cval{0.99}0.01 \\
         Mistral 7B$_{G}$ & \cval{0.18}0.41 & \cval{0.68}1.18 & \cval{0.99}0.01 \\
         Gemma 7B & \cval{0.78}0.09 & \cval{0.38}1.53 & \cval{0.98}0.02 \\
         Gemma 7B$_{G}$ & \cval{0.87}0.05 & \cval{0.22}1.75 & \cval{1.00}0.00 \\
         Qwen 7B & \cval{0.00}\textbf{0.59} & \cval{1.00}0.41 & \cval{0.99}0.01 \\
         Qwen 7B$_{G}$ & \cval{0.03}0.57 & \cval{0.98}0.45 & \cval{0.99}0.01 \\
         \hline
         Therapist & \cval{0.68}0.18 & \cval{0.65}0.91 & \cval{0.00}\textbf{0.19} \\
         \hline
    \end{tabular}
    }
    \caption{Empathy evaluation.}
    \label{tab:empathy}
\end{table*}

\begin{table*}[h!]
\centering
\small
\setlength{\tabcolsep}{6pt}
\renewcommand{\arraystretch}{1.2}

\begin{tabular}{p{0.47\linewidth} p{0.47\linewidth}}
\toprule
\textbf{Session A – Conversation History} & \textbf{Session B – Conversation History} \\
\midrule

\includegraphics[width=0.05\linewidth]{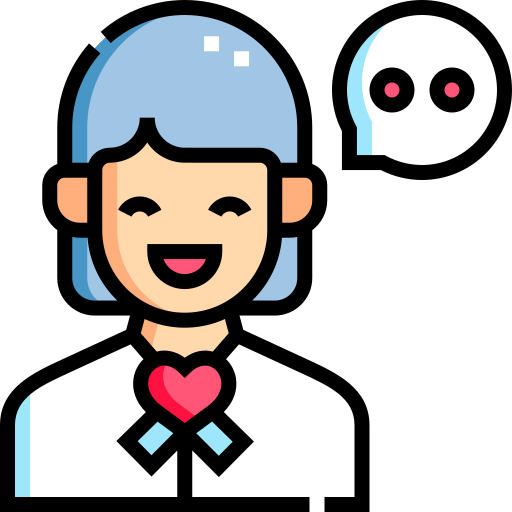}
\textbf{Therapist:} Did you really eat nothing at all? \newline
\includegraphics[width=0.05\linewidth]{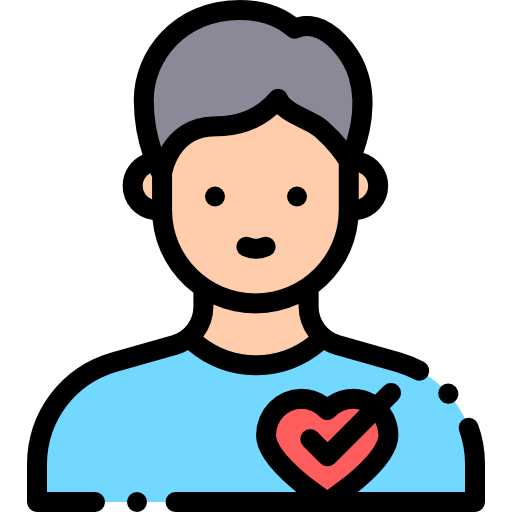}
\textbf{Client:} Yeah. \newline
\includegraphics[width=0.05\linewidth]{figures/speech_9689229.png}
\textbf{Therapist:} Can you tell me more about why that day was challenging? Was there anything happening around that time? \newline
\includegraphics[width=0.05\linewidth]{figures/love_756736.png}
\textbf{Client:} It has been quite busy at school because we have our exams coming up, and we even had a test scheduled for the following day. I was feeling a bit worried about it, so I tried to focus on studying instead.

&
\includegraphics[width=0.05\linewidth]{figures/speech_9689229.png}
\textbf{Therapist:} His sense of hopelessness is rooted in a lot of reality. \newline
\includegraphics[width=0.05\linewidth]{figures/love_756736.png}
\textbf{Client:} Yes, exactly. \newline
\includegraphics[width=0.05\linewidth]{figures/speech_9689229.png}
\textbf{Therapist:} While you're doing this work daily, do you think there's any hope at all? \newline
\includegraphics[width=0.05\linewidth]{figures/love_756736.png}
\textbf{Client:}Certainly. I have faith that there is always some form of hope. I believe in resilience. I think once you tap into it, you can accomplish almost anything. However, it can be challenging to navigate through the various layers that define this generation of children. It can be exhausting.

\\

\midrule
\multicolumn{2}{c}{\textbf{Expert and Generated Therapist Replies}} \\
\midrule

\includegraphics[width=0.05\linewidth]{figures/speech_9689229.png}
\textbf{Therapist:} Were you feeling anxious about the upcoming exam? Was there anything else worrying you?

&
\includegraphics[width=0.05\linewidth]{figures/speech_9689229.png}
\textbf{Therapist:} It's difficult to get through.
\\[4pt]

\includegraphics[width=0.05\linewidth]{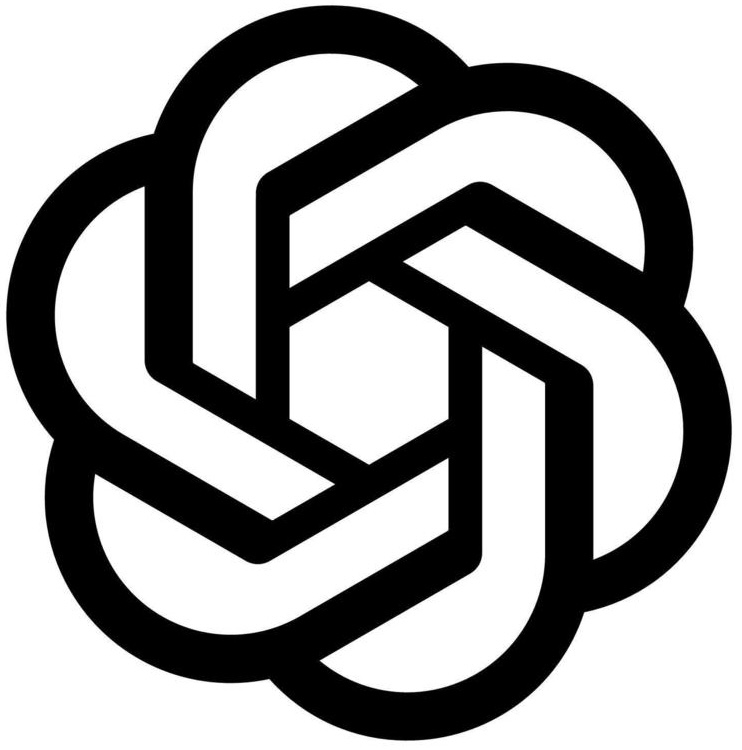}
\textbf{GPT-4o-mini$_G$:} It makes sense that preparing for exams could create a lot of stress. How do you typically manage that worry, and what strategies could help you feel more balanced during busy times like this?

&
\includegraphics[width=0.05\linewidth]{figures/gpt.jpg}
\textbf{GPT-4o-mini$_G$:} It's powerful that you believe in resilience and hope despite the challenges. What strategies or approaches have you found effective in helping kids tap into that resilience?
\\[4pt]

\includegraphics[width=0.05\linewidth]{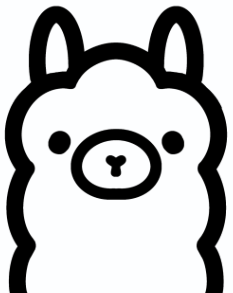}
\textbf{LLaMA3 8B$_G$:} Can you tell me more about what was going through your mind during that test, and how it affected your mood on that day?

&
\includegraphics[width=0.05\linewidth]{figures/llama.png}
\textbf{LLaMA3 8B$_G$:} As I reflect on this situation and its relation to my own sense of hope, I'm struck by the resilience you mentioned earlier. It's remarkable how you continue to believe in the potential for positive change despite the challenges you're facing with this generation of kids. Can you tell me more about what drives your passion for this work, even when it feels like an uphill battle?
\\[32mm]

\includegraphics[width=0.05\linewidth]{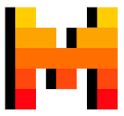}
\textbf{Mistral 7B$_G$:} Can you identify the key issues that might make it difficult for you to talk to your father this week? We can work together on strategies to address these challenges.

&
\includegraphics[width=0.05\linewidth]{figures/mistral.png}
\textbf{Mistral 7B$_G$:}  I appreciate your insight about the challenges you face with this generation of kids. It's important to recognize the resilience within them and ourselves. Let's focus on finding ways to tap into that resilience effectively. Can we explore some strategies for maintaining hope while addressing these layers in a manageable way?
\\[30mm]

\includegraphics[width=0.05\linewidth]{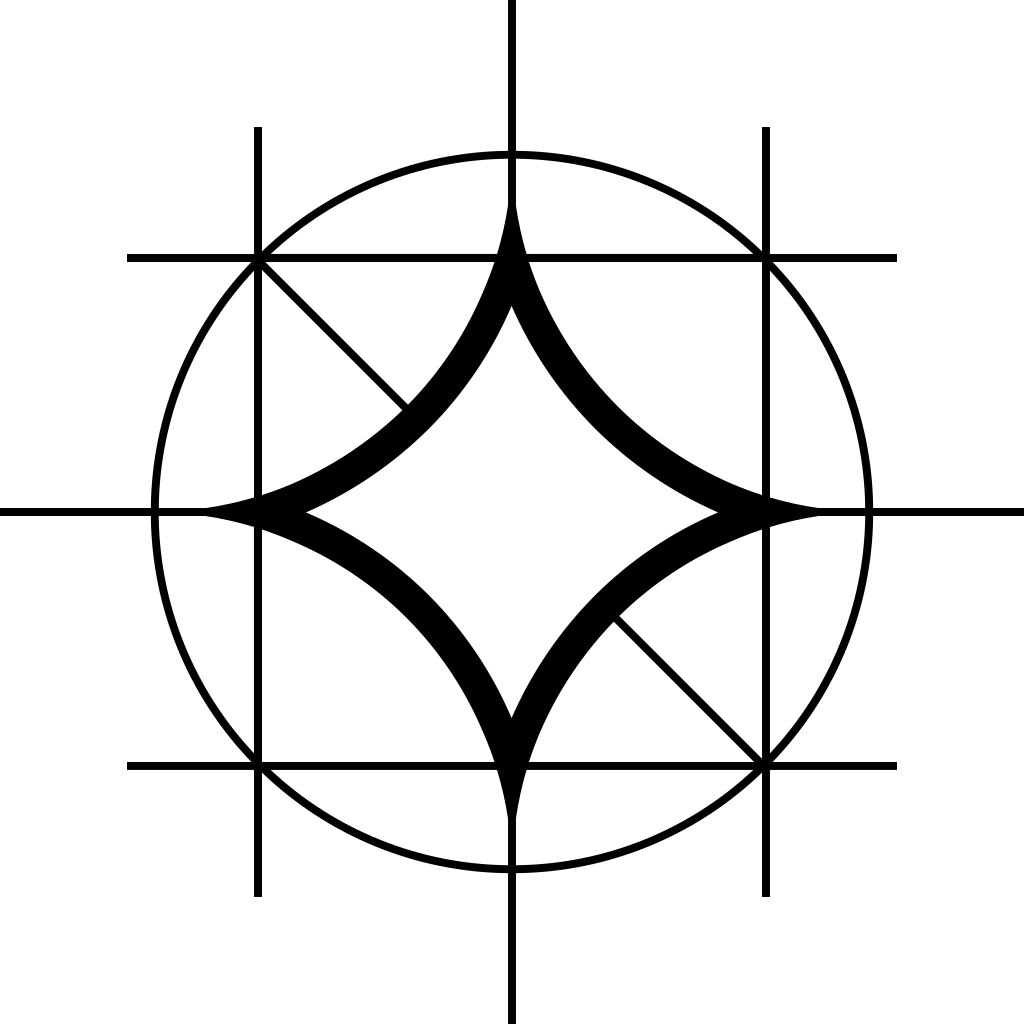}
\textbf{Gemma 7B$_G$:} It sounds like the exams were really weighing on you.  Can you tell me more about how that was making you feel?

&
\includegraphics[width=0.05\linewidth]{figures/gemma-black.png}
\textbf{Gemma 7B$_G$:} What makes you feel hopeful despite those challenges?
\\[4pt]

\includegraphics[width=0.05\linewidth]{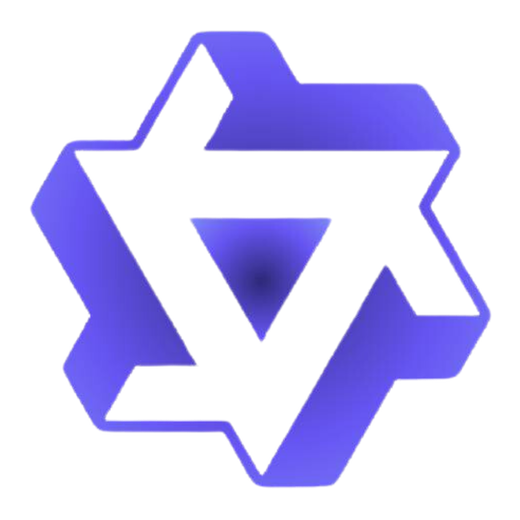}
\textbf{Qwen 7B$_G$:} That makes sense. That makes sense. It sounds like you were feeling overwhelmed by both your upcoming exam and the potential for conflict with your family during the holiday season. Can you tell me more about how you're planning to manage these emotions? Have you tried any coping strategies in the past?

&
\includegraphics[width=0.05\linewidth]{figures/qwen.png}
\textbf{Qwen 7B$_G$:} It's clear that you have been doing a lot of self-reflection and work on resilience. That's an excellent start. Regarding identifying your negative core beliefs, it's important to proceed at a pace that feels comfortable for you. If trying to evaluate these beliefs too early causes distress, we can certainly adjust our approach. [...] Remember, therapy is about taking small steps and making progress over time. Let's continue this process together.
\\

\bottomrule
\end{tabular}

\caption{Conversation samples from two different CBT sessions and generated replies from the LLMs.}
\label{tab:conv-sample}
\end{table*}

In this section, we present the results of our evaluation of the proprietary and open-source models for linguistic quality, semantic coherence, and therapeutic fidelity using standard natural language generation metrics, natural language inference metrics, and automated scoring for general and CBT-specific skills assessment.

In Table \ref{tab:text-generation-metrics}, we present the results in terms of BLEU, ROUGE, METEOR, BERTScore, Distinct-1, and natural language inference metrics, Mean Consistency, and Maximum Contradictions. In terms of linguistic similarity and semantic alignment, GPT-4o-mini and its RAG version achieve the strongest overall performance in terms of lexical overlap and semantic similarity, most closely resembling human therapists. GPT-4o-mini$_{G}$ achieves the highest scores in terms of ROUGE-1 (17.05), and BERTScore (85.82), and the second-highest METEOR score (17.10). 

The small improvement offered by the RAG approach suggests that these models already encode substantial CBT-related knowledge, aligning with prior findings \cite{zhang2024cbt}. The low BLEU scores indicate that the text generated by the language models has a limited word overlap with the therapists' responses used as references. Despite this, all models achieved high BERTScores, indicating their general ability to capture the meaning of therapist responses. However, we observe the greatest differences in consistency, contradiction, and diversity: Qwen 7B demonstrates the highest logical consistency with therapist intent, shown in the high score in Mean Consistency (95.17), whereas Gemma 7B$_{G}$ achieved the lowest contradiction score (17.46) and great linguistic diversity (97.67).

In Table \ref{tab:llm-eval}, we present the results of the evaluation of general counseling and CBT-specific skills. We present results for general counseling skills, including understanding, interpersonal effectiveness, and collaboration, as well as CBT-specific skills, such as guided discovery, focus, and strategy. All models underperform relative to the human therapist baseline, particularly in areas requiring adaptive reasoning and nuanced judgment. Regarding general counseling skills, human therapists demonstrate a better understanding of their clients' issues and concerns. They are more successful in maintaining a positive therapeutic relationship (interpersonal effectiveness) and in engaging clients collaboratively in goal-setting and decision-making. Although some models replicate the structural elements of CBT dialogue, they lack the contextual sensitivity needed for effective therapeutic engagement. Human therapists tend to score higher in identifying and addressing client behaviors that require improvement (focus), as well as in following strategies to promote change in these problematic behaviors (strategy). Notably, the only area in which the LLMs scored higher in therapy skills is guided discovery, suggesting that these models attempt to help clients engage in self-reflection.

In Table \ref{tab:empathy}, we present the results of the empathy evaluation, in terms of empathetic reactions, explorations, and interpretations. Most models score highest in exploration, frequently asking follow-up questions to probe clients' emotions. However, this behavior often becomes excessive, reflecting a procedural rather than emotionally attuned form of empathy. LLama3 8B$_{G}$ scored the highest (1.96) in terms of explorations, with 2 being the maximum possible score, suggesting that the model almost always tries to ask follow-up questions, and explore the client's feelings, rather than using other therapeutic strategies. Models such as Mistral 7B and Qwen 7B show greater emotional expressiveness through empathic reactions. However, even these responses often rely on formulaic phrasing instead of demonstrating genuine affective or cognitive understanding. All LLMs receive very low scores in one of the most important aspects of empathy: interpretations. This aspect is related to cognitive empathy, which refers to the therapist's ability to understand the client's emotions. While LLMs seem to excel at simulating affective empathy through empathetic reactions, they show limited evidence of cognitive empathetic responses. 

Based on the results presented in this section, we can respond to our research questions:

\paragraph{RQ1} Our results indicate that LLMs can emulate professional therapists, with GPT-4o-mini and its retrieval-augmented variant achieving the highest ROUGE, METEOR, and BERTScore values. Qwen 7B demonstrated strong logical consistency while Gemma 7BG achieved the lowest contradiction score, indicating coherent dialogue generation aligned with the therapist's intent. However, the models’ tendency to produce longer, more elaborate responses than human therapists suggests that, while linguistically fluent, they may lack the brevity and precision typical of authentic exchanges during therapy sessions.

\paragraph{RQ2} Our findings suggest that RAG provides limited improvement over generation-only approaches. Although the RAG variants of some models, such as GPT-4o-mini$_{G}$ and Gemma 7B$_{G}$, achieved slightly higher scores on some text generation metrics, these gains were minimal. The results suggest that LLMs already possess sufficient CBT-related knowledge, enabling them to generate contextually coherent and semantically aligned responses without requiring external information.

\paragraph{RQ3} The human therapist baseline consistently outperformed all models in terms of general and CBT-specific skills. Regarding empathy, even if some LLMs scored high in explorations, suggesting that the model almost always attempts to ask follow-up questions, this approach is not effective in a therapeutic context. All models received low scores in cognitive empathy, especially in their interpretations, highlighting a gap in their understanding of client emotions, despite their ability to simulate affective empathy.

\section{Qualitative analysis}

In the previous section, we presented only quantitative results regarding the effectiveness of LLMs for CBT sessions. In this section, we will also provide a qualitative analysis of the generated responses. In Table \ref{tab:conv-sample}, we present two examples of conversations from two different therapy sessions. One noticeable pattern in the data is that while the therapists' original utterances are short, the language models tend to generate more detailed and longer responses.

\subsection{Agreeableness Bias}

A manual inspection of the results reveals a pattern of agreeableness bias across several models, consistent with findings by \citet{salecha2024large} that large language models tend to produce overly affirming responses. Examples from GPT-4o-mini (“Absolutely, your feelings and decisions should be what matter most”, ''Absolutely, it's important to recognize that your relationship should be defined by what works for both of you, not by arbitrary standards.'') and Qwen 7B (“Absolutely. It's important to base our thoughts and beliefs on reality rather than just what we fear might be true”) illustrate how the models often respond with enthusiastic validation rather than critical exploration. Similarly, phrases such as “It's completely understandable to feel that way,” frequently produced by Mistral 7B, Qwen 7B, and GPT-4o-mini variants, demonstrate a tendency toward reassurance. While these responses convey warmth and support, they reflect surface-level empathy that reinforces rather than explores the client’s perspective \cite{curry2023computer}. Always agreeing with clients' views does not encourage them to examine and reframe their thoughts, which is an important part of CBT \cite{sharma2023cognitive}.

\begin{CJK*}{UTF8}{gbsn}

\subsection{Code-switching}

Another issue in response generation is that Qwen 7B sometimes produced bilingual output (e.g., partial Chinese sentences), indicating inconsistencies in encoding: ``Remember that简化信息有助于我们更好地理解和处理问题。'' (En: ``Remember that simplifying information helps us better understand and deal with problems.''), ``In the next session, let's practice some cognitive restructuring techniques where you'll be able to reframe your担忧 into more adaptive ones.'' (En: ``In the next session, let's practice some cognitive restructuring techniques where you'll be able to reframe your concerns into more adaptive ones.'').
\end{CJK*}

\section{Conclusion and Future Work}
In this paper, we show that LLMs have the potential to replicate certain elements of Cognitive Behavioral Therapy, particularly in terms of linguistic coherence and supportive tone. However, they have notable limitations in general therapeutic skills, CBT-specific skills, and empathy. Retrieval-augmented generation shows limited improvement, suggesting that LLMs possess sufficient CBT-related knowledge to generate contextually coherent, semantically aligned responses without the need for external information. These results indicate that while LLMs show promising linguistic and structural competence in simulating CBT and mimicking the language of a CBT therapist, they do not replicate the nuanced therapeutic skills of human therapists that are essential for achieving positive therapeutic outcomes. In future research, we aim to investigate LLMs' ability to respond to clients and potential users from diverse cultural backgrounds. Previous studies have shown that emotional nuances can differ across populations in therapeutic settings \cite{abbasi2025hamraz}, and that the expression of mental health distress can also vary across different cultures \cite{bucur2025survey}.

\section*{Limitations}
This study has several limitations that should be acknowledged. First, the dataset used in our evaluation is relatively small, consisting of a limited number of role-played CBT sessions. Although this dataset provides valuable initial insights, it does not capture the full diversity of therapeutic contexts or client presentations. In future work, we plan to expand our dataset by collecting a larger and more varied sample of authentic and simulated therapy dialogues to enhance the robustness and generalizability of our findings.

Second, while we employed automated evaluation methods validated in prior research for assessing therapeutic skills \cite{lee2024cactus} and empathy \cite{sharma2020computational}, these approaches cannot fully replace manual evaluation by expert CBT practitioners. Expert human judgment is necessary to capture the subtle nuances of empathy and therapeutic reasoning that automated metrics may overlook. Therefore, future studies should incorporate expert evaluations to supplement quantitative analyses and provide a more comprehensive assessment of model performance.

Lastly, we evaluated the performance of small-scale language models, specifically those with 7-8B parameters, and GPT-4o-mini, due to our limited computational resources. Our findings suggest that while these models can generate contextually coherent responses that resemble CBT, they still do not match the skills demonstrated by human therapists. In future work, we plan to assess larger models to determine if the results are consistent.

Despite these limitations, we believe this paper represents an important first step toward systematically evaluating LLMs in CBT contexts. By combining linguistic, semantic, and therapeutic dimensions, our work lays the groundwork for large-scale assessments of AI-assisted psychotherapy.

\section*{Ethical Considerations}
All data were derived from publicly available, non-clinical role-play sessions. This research does not endorse the use of LLMs in psychotherapy. We emphasize that LLM-based counseling tools must include clear disclaimers, human oversight, and rigorous safety validation before being deployed, due to risks such as hallucination, reinforcement of cognitive distortions, and inappropriate emotional responses. 

\section*{Bibliographical References}
\bibliography{main}
\bibliographystyle{lrec2026-natbib}

\end{document}